# Optimal Design and Analytical Modeling of a Soft Fin-Ray Effect Gripper Finger Using the Finite Rigid Elements Method

S. Adeli, H. Sayyaadi

*Abstract*— **Fin Ray-inspired soft grippers offer a promising solution for gently handling delicate, irregular objects, especially in agriculture. The objective of this research is to design, fabricate, and model a Fin Ray Effect (FRE) soft gripper finger to enable precise force control in future applications. This design aims to gently grasp delicate agricultural products, such as tomatoes, that require both adaptability and accurate force application. To address the inherent challenges of soft robotics, including nonlinear behavior, infinite degrees of freedom, and variable material properties, the Finite Rigid Elements Method (FREM) was employed for modeling. This method preserves analytical accuracy while providing a reliable foundation for the development of a force controller in later stages. A detailed Finite Element Model (FEM) was created using ANSYS, and the analytical results were validated through simulation and experimental testing. The gripper's fingers were optimized based on four key criteria: tip displacement, total deflection, stress distribution, and contact force. The optimal finger configuration includes a length of 30 mm, rib spacing of 10 mm, seven ribs angled at -15°, and a rib thickness of 1 mm. Theoretical modeling using the FREM predicted finger deformation with a 3% error, while the ANSYS numerical model achieved 2% error.**

## I. INTRODUCTION

SOFT robotic grippers provide adaptable and safe manipulation of delicate objects, addressing limitations of rigid robotic systems. The fin ray effect gripper, inspired by fish fin mechanics, combines structural simplicity with effective grasping of irregular shapes. These features make it well-suited for agricultural applications, where gentle and reliable crop handling is critical. Fish fin-inspired gripper fingers have garnered significant attention in the field of soft robotics due to their exceptional ability to conform to and securely grasp objects of varying shapes, sizes, and materials. The performance of these grippers is highly dependent on several key design parameters, including the choice of material, finger length, width, depth, wall thickness, number and spacing of ribs, rib angle, and rib thickness. By carefully tuning these parameters, the adaptability, compliance, and gripping strength of the fingers can be optimized to handle a wide range of geometries.

Finite element simulations are widely used to design and optimize fish fin–inspired gripper fingers, typically through two methods: displacing a circular object toward the finger to assess conformity, or applying distributed forces to evaluate structural response under realistic loading. Key performance metrics include fingertip displacement, contact force, winding number, and maximum deflection. Displacement in the X direction indicates compliance, while Y-direction displacement reflects finger closure; higher contact force denotes stiffness, greater winding number suggests secure gripping, and larger deflection indicates adaptability with less applied force. Additional parameters such as stress, strain, and normalized contact area further inform design optimization: lower stress enhances durability, higher strain improves flexibility, and larger contact area ensures stable interaction. Together, these criteria provide a comprehensive framework for evaluating and improving the mechanical performance of gripper fingers in diverse applications.

Research highlights that both material selection and structural design are critical to the performance of soft robotic grippers. Materials such as TPU and Ninjaflex are widely used for their flexibility, durability, and resilience under repeated deformation [1][2][3]. Geometric factors, including rib structure, spacing, thickness, and angle, directly influence adaptability, stiffness, and force distribution. Studies show that optimizing rib spacing and thickness improves the balance between flexibility and strength [2], while angled ribs enhance deformation behavior and grip security [4][5][6][7]. Simplified designs without fillers provide high flexibility but limited stiffness, whereas cross-linked or branched internal structures achieve better force output [5][8]. Additional design features such as optimized rib numbers, wall thicknesses, and external wall angles have demonstrated improved uniform force distribution, adaptability to irregular shapes, and safe handling of delicate crops like apples [9][10]. Multi-finger and Fin Ray–based designs with curved or variable-angle ribs further enhance compliance, reduce stress concentration, and enable reliable manipulation of fragile or irregular objects in agricultural and industrial applications [11][12][13].

System dynamics identification and robot modeling play a central role in both optimizing robotic design and enabling intelligent behavior. Modeling approaches, including finite element analysis, numerical simulations, and machine learning,

S. Adeli is with the Mechanical Engineering Department at Sharif University of Technology, Tehran, Iran (e-mail: saraadeli0215@gmail.com).

H. Sayyaadi is with the Mechanical Engineering Department at Sharif University of Technology, Tehran, Iran (e-mail: sayyaadi@sharif.edu).

provide valuable tools for deriving optimal designs and achieving reliable performance. Research on fish fin-inspired soft gripper's modeling techniques generally falls into two domains: numerical and data-driven modeling of flexible bodies, and theoretical frameworks for deriving governing equations of soft actuators. Advancing these directions enables more accurate control strategies and practical deployment of soft grippers in sensitive agricultural tasks.

Recent research has increasingly explored machine learning and data-driven approaches to model Fin Ray–inspired soft fingers. Recent Research proposed a deep learning framework that predicts contact forces and stress distribution using a CNN network trained by FEM-generated images, offering significant speed improvements over traditional simulations for real-time applications [14]. Physics-informed neural networks (PINNs), such as PINN-Ray [15], integrate elastic mechanics principles with experimental data to accurately simulate complex deformations while reducing computational cost. Similarly, hybrid physics-based recurrent neural networks have been applied to pneumatic soft actuators, combining data-driven modeling with physical laws to predict dynamic responses efficiently [16][17]. Zhang et al. [18] analyzed the lateral motion of a pneumatically actuated soft fish-like robot using neural networks and experimental data, though requiring extensive retraining for structural or environmental changes. Progressive learning strategies have also been used to continuously update neural network models, enhancing prediction accuracy and control precision for soft actuators [19]. Together, these studies demonstrate the potential of combining machine learning with physics-based modeling for efficient and reliable soft robotic modeling. Despite their advantages, these methods often require large training datasets, significant computational resources, and frequent retraining when system parameters or environments change. Moreover, their black-box nature can limit interpretability and robustness, making them less practical for reliable, real-time control in dynamic soft robotic applications.

The second category of modeling literature focuses on theoretical approaches to analyze Fin Ray–inspired soft fingers. Several studies have employed kinematic-static and kinetostatic models, often based on modified chain beam constraints or discrete flexible beam methods, to predict finger deformation, contact forces, and adaptive efficiency [20][21][22][23]. Cassrat rod theory has been widely used to model the nonlinear dynamics of soft beams, capturing large deformations, bending, torsion, and tension under various loading conditions [21][23][24][25]. These models are typically validated through finite element simulations and experimental tests, demonstrating accurate predictions of finger kinematics, reaction forces, and structural stability. Additional approaches include modular closed-chain models for multi-finger grippers [26], 3D deformation frameworks for soft actuators [27], and discretized bending actuators analyzed via torsional spring systems and Denavit-Hartenberg parameters [28]. While highly accurate, these methods are computationally complex, limiting their direct use for real-time control.

The objective of this research is to design, fabricate, and model a fish fin–inspired gripper finger for precise force control in handling delicate agricultural products, such as tomatoes. The finger, 3D-printed from TPU, offers high flexibility, strength, and reduced fabrication time. Challenges arise from TPU's nonlinear behavior, variable Young's modulus, and the infinite degrees of freedom in soft robots. To address these, the finite rigid elements method was applied to achieve accurate yet computationally efficient modeling suitable for controller design. A finite element model in ANSYS further supported performance evaluation, and theoretical results were validated experimentally.

## II. Material and Methods

### *A. Design and Fabrication of a Fin Ray-Effect Finger*

In this study, the gripper finger was modeled in ANSYS Workbench to evaluate its mechanical performance. The simulation involved displacing a spherical object toward the middle region of the finger, allowing assessment of the design's adaptability and flexibility when interacting with delicate agricultural products, such as tomatoes. For simulation, a spherical object with a diameter of 60 mm was considered to represent a tomato. This object, with the physical properties defined in TABLE I, is in contact with the middle section of the gripper's finger, specifically, the 50 mm central length, located approximately 25 mm from the fingertip. The finger is made of TPU 95A, a material selected for its high flexibility and superior mechanical strength compared to alternatives like silicone. The physical properties of TPU 95A are listed in TABLE II. In this simulation, the spherical object is displaced 20 mm perpendicular to the finger surface. The contact simulation in ANSYS is illustrated in Fig. 1.

Table I
Physical properties of tomatoes [29].

| DENSITY (KG/M3) | YOUNG'S MODULUS (PA) | POISSON'S RATIO | BULK MODULUS (PA) | SHEAR MODULUS (PA) |
|---|---|---|---|---|
| 470 | $1.1\times10^{8}$ | -0.0678 | $3.23\times10^{7}$ | $5.9\times10^{4}$ |

Table II
Physical properties of TPU 95A [13].

| YOUNG'S MODULUS (PA) | POISSON'S RATIO | BULK MODULUS (PA) | SHEAR MODULUS (PA) | YIELD STRESS (PA) |
|---|---|---|---|---|
| $2.6\times10^{7}$ | 0.481 | $2.28\times10^{8}$ | $8.78\times10^{6}$ | $8.6\times10^{6}$ |

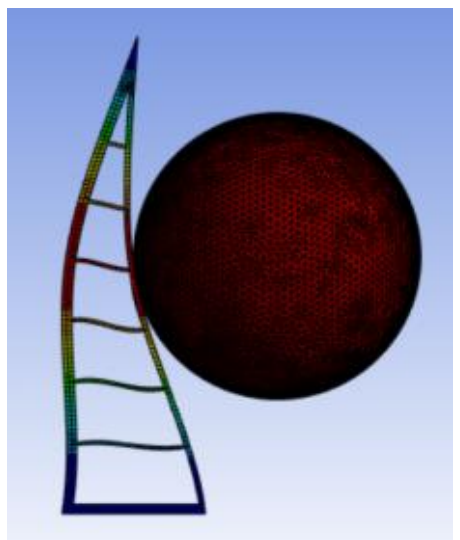

**Fig. 1.** Bending induced in the finger due to the displacement of the spherical object toward the finger.

The optimal design of the gripper finger aims to enhance its performance in grasping delicate objects, particularly agricultural products. This was achieved by analyzing the displacement of a spherical object toward the finger using FEM within the Static Structural module of ANSYS Workbench. The optimal performance of the Fin Ray–effect gripper finger depends on carefully selected design parameters governing stiffness, flexibility, and adaptability. Key factors include material properties, which dictate strength and stress response; overall dimensions (length, width, height); beam wall thicknesses, which influence stiffness and force transmission; and rib geometry (number, spacing, thickness, angle), which controls force distribution and object conformity. The identified design parameters are illustrated in Fig. 2.

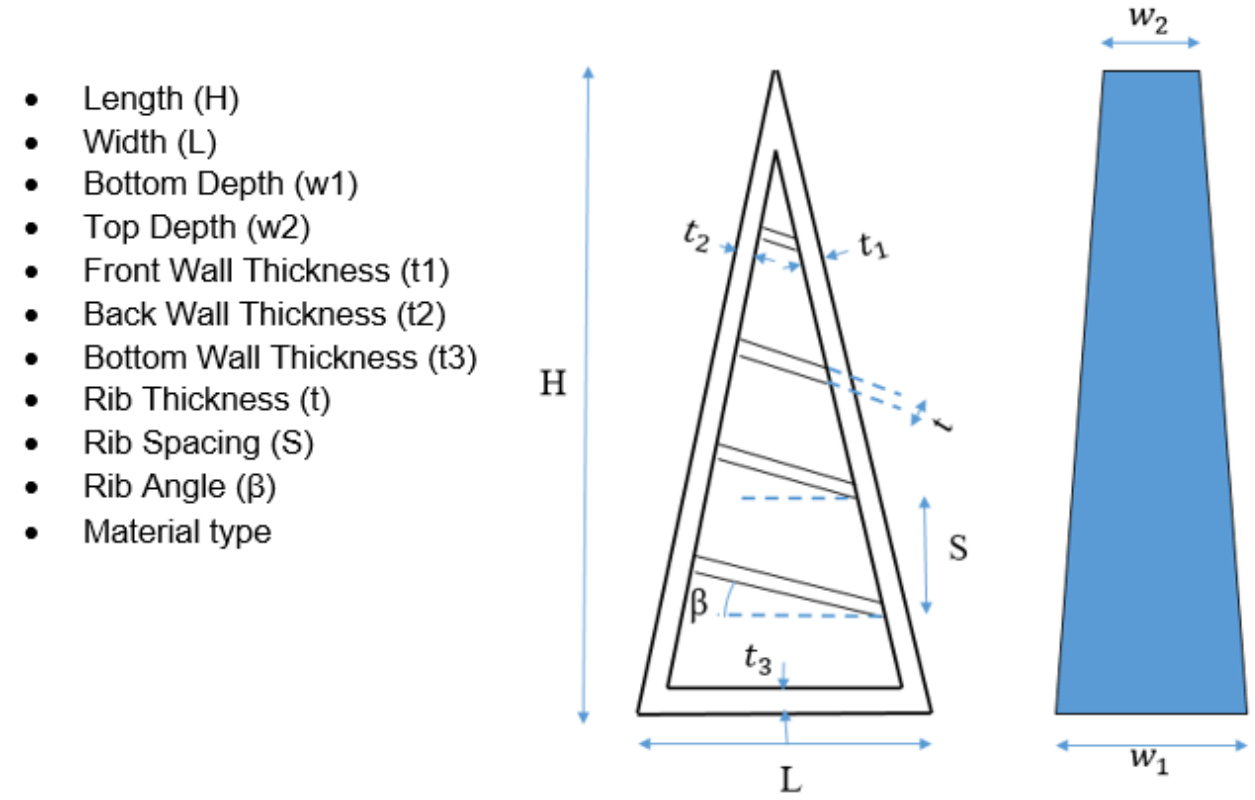

**Fig. 2.** Design Parameters of the Fin Ray-Effect-Based Soft Gripper Finger.

The overall structure of the gripper finger was designed with a trapezoidal longitudinal beam, featuring a base width of 32 mm, a tip width of 16 mm, and a height of 100 mm (Fig. 3 (b)). This geometry improves the finger's ability to grasp objects effectively while ensuring sufficient strength. Based on previous studies [1], a stiffer back beam enhances positive contact forces, allowing the finger to conform more tightly to the object's surface and minimizing lateral deformation. Accordingly, the front beam thickness was set to 1 mm, while the back and bottom beams were set to 2 mm to maintain overall structural stiffness and integrity (Fig. (a)).

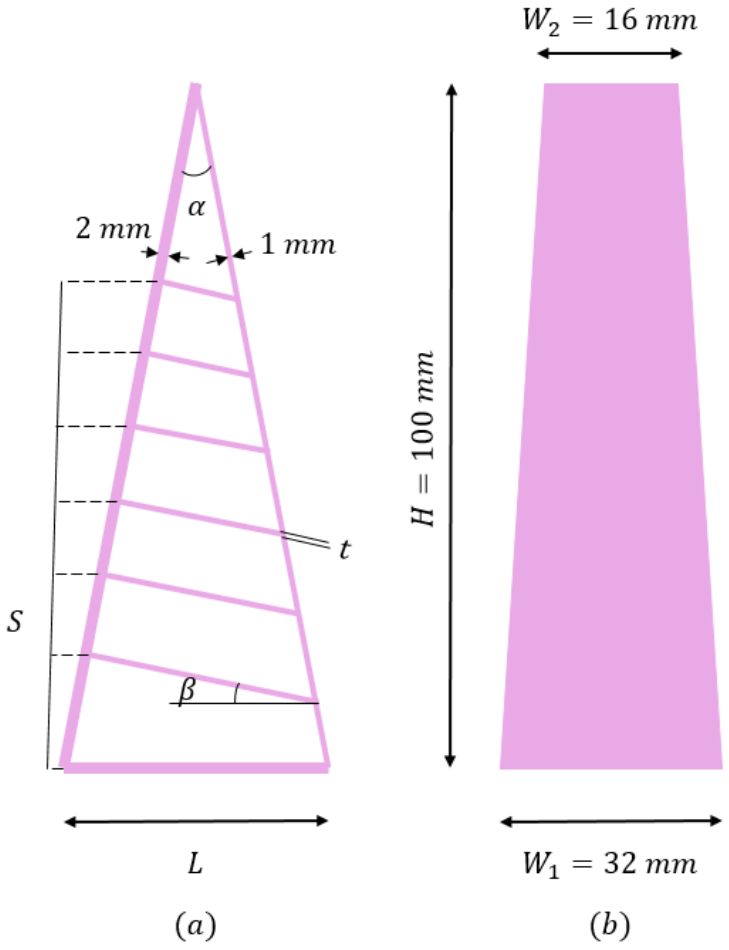

**Fig. 3.** Parametric Model of the Fin Ray Finger.

To systematically investigate the influence of geometric parameters, four key variables, finger width (L), rib spacing (S), rib angle (β), and rib thickness (t), were varied across defined ranges. Width was varied from 30 to 60 mm in 10 mm increments, rib spacing from 6 to 18 mm in 2 mm steps, rib angle from –15° to 15° in 5° increments, and rib thickness from 0.4 to 1.8 mm in 0.2 mm steps. These variations allowed a comprehensive parametric analysis, evaluating the effect of each parameter on force distribution, stiffness, and adaptability of the finger. To evaluate stiffness and flexibility, four key metrics were considered: fingertip displacement in the X and Y directions, resultant contact force, overall deflection, and internal stress. X-direction displacement reflects adaptability to object position changes, Y-direction displacement indicates enclosure capability, resultant contact force measures load-bearing capacity, and stress evaluates structural durability.

*B. Dynamic Modeling of the Fin Ray-Effect Finger*

The dynamic modeling of the Fin Ray–effect inspired soft gripper is developed using a structured approach to provide accurate and reliable system dynamics while maintaining analytical simplicity. Several assumptions are made to reduce complexity: the flexible section of the finger undergoes only bending, with torsion and longitudinal stretching neglected; motion is confined to a two-dimensional plane along the curvature of the neutral axis; external forces are applied perpendicularly to a central line representing the flexible segment, contributing directly to bending; and only serial connections between rigid elements are considered, ignoring parallel interactions. Based on these assumptions, the gripper body is divided into rigid elements, and dynamic relationships between them are defined. Planar motion is simulated by connecting rigid segments with torsional springs and parallel dampers, representing the bending and damping characteristics of the finger. The schematic of the modeled system compared to the actual system is shown in Fig. 4.

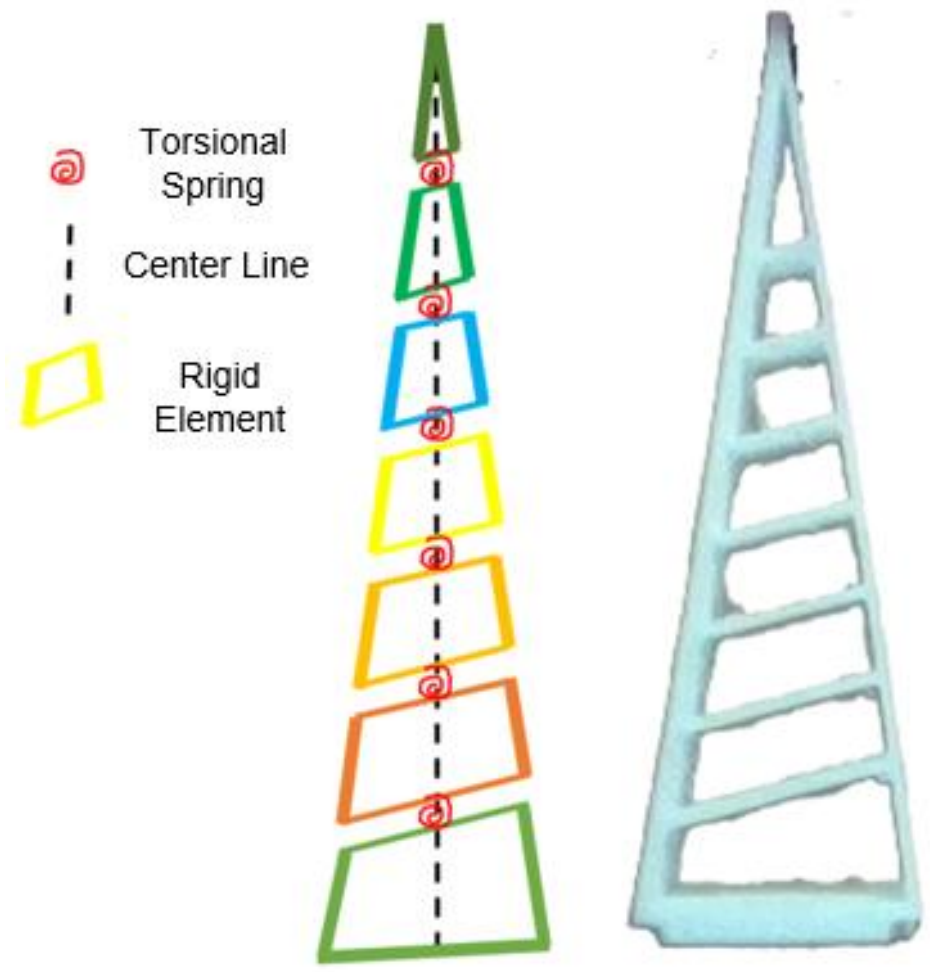

**Fig. 4.** Schematic of the Fin Ray Effect Finger.

The gripper is considered stationary with no acceleration, and the system is modeled with 8 degrees of freedom aligned along the symmetry axis, with local coordinates defined using the Craig method, as illustrated in Fig. 5.

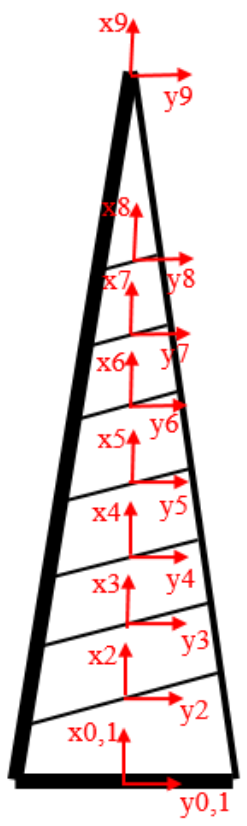

**Fig. 5.** Schematic of Local Frame Assignment for Calculating Denavit-Hartenberg Parameters.

Using Denavit-Hartenberg parameters, joint relationships and motions are systematically recorded, and dynamic equations are derived in joint space via the Newton-Euler method. Internal torques from dynamic coupling and external torques due to contact forces, including normal and friction forces, are calculated. These results are integrated into comprehensive governing dynamic equations, providing a foundation for simulation, analysis, and control of the gripper's performance.

Considering the dimensions of the finger and the coordinate frames defined in Fig. 5, the Denavit-Hartenberg table is obtained as shown in TABLE III.

Table III
DENAVIT-HARTENBERG PARAMETER TABLE.

| i | $a_{i-1}[mm]$ | $\alpha_{i-1}[°]$ | $\theta_i[°]$ | $d_i[mm]$ |
|---|---|---|---|---|
| **1** | 0 | 0 | $\theta_1$ | 0 |
| **2** | 13 | 0 | $\theta_2$ | 0 |
| **3** | 10 | 0 | $\theta_3$ | 0 |
| **4** | 10 | 0 | $\theta_4$ | 0 |
| **5** | 10 | 0 | $\theta_5$ | 0 |
| **6** | 10 | 0 | $\theta_6$ | 0 |
| **7** | 10 | 0 | $\theta_7$ | 0 |
| **8** | 10 | 0 | $\theta_8$ | 0 |
| **9** | 29 | 0 | 0 | 0 |

Using the parameters extracted from the DH table, the homogeneous transformation matrix $T_i^{i-1}$ with dimensions 4×4 can be calculated according to (1) [30]. In this equation, the terms $\cos\theta_i$ and $\sin\theta_i$ are respectively abbreviated as $c\theta_i$ and $s\theta_i$.

$$ {}^{i-1}T_{i[4\times4]} = \begin{bmatrix} c\theta_i & -s\theta_i & 0 & a_{i-1} \\ s\theta_i c\alpha_{i-1} & c\theta_i c\alpha_{i-1} & -s\alpha_{i-1} & -s\alpha_{i-1} d_i \\ s\theta_i s\alpha_{i-1} & c\theta_i s\alpha_{i-1} & c\alpha_{i-1} & c\alpha_{i-1} d_i \\ 0 & 0 & 0 & 1 \end{bmatrix} \tag{1} $$

This transformation matrix consists of $R_i^{i-1}$, a 3×3 rotation matrix, and $p_i^{i-1}$, a 3×1 translation vector, as shown in (2). The rotation matrix defines the orientation of frame i relative to frame i-1, while the translation vector represents the displacement of the origin of frame i with respect to frame i-1.

$$ {}^{i-1}T_{i[4\times4]} = \begin{bmatrix} R_{(i-1)}^{(i)} & p_{(i-1)}^{(i)} \\ 0_{(1\times3)} & 1_{(1\times1)} \end{bmatrix} \tag{2} $$

According to (3), the homogeneous transformation matrix of the center of mass of each link can be expressed relative to the inertial frame [30].

$$ T_i^{(0)} = T_1^{(0)} \times T_2^{(1)} \times T_3^{(2)} \times \cdots \times T_i^{(i-1)} \tag{3} $$

The Newton-Euler method is used to derive the governing dynamic equations of the system. In this approach, the effects of mass and inertia of the rigid components for each degree of freedom are first modeled. To this end, the distribution of angular velocity is considered, moving from the first degree of freedom towards higher degrees, according to (4) [30].

$$ {}^{i+1}\boldsymbol{\omega}_{i+1[3\times1]} = {}^{i+1}\mathbf{R}_{i[3\times3]}\, {}^{i}\boldsymbol{\omega}_{i[3\times1]} + \dot{\theta}_{i+1[1\times1]}\, {}^{i+1}\mathbf{z}_{i+1[3\times1]} \tag{4} $$

In this relation, ${}^{i+1}\boldsymbol{\omega}_{i+1}$ represents the angular velocity vector of link i+1 with respect to the inertial frame, expressed in the coordinate frame i+1, and is a 3×1 vector. ${}^{i+1}\mathbf{R}_i$ is the 3×3 rotation matrix that transforms vectors from frame i to frame i+1. $\dot{\theta}_{i+1}$ denotes the angular velocity of link i+1 about its own z-axis, and ${}^{i+1}\hat{\mathbf{z}}_{i+1}$ is the unit vector along the z-axis of frame i+1, also represented as a 3×1 vector. For calculating the rotation matrix, according to (2), since the rotation matrix is orthonormal, it can be computed using (5) [30].

$$ {}^{i+1}\mathbf{R}_i = {}^{i-1}\mathbf{R}_i^{\mathrm{T}} \tag{5} $$

To propagate the angular acceleration, moving from the first link toward the last link, the values are calculated using the terms from (2), according to (6) [30].

$$ {}^{i+1}\dot{\boldsymbol{\omega}}_{i+1[3\times1]} = {}^{i+1}\mathbf{R}_i \left( {}^{i}\dot{\boldsymbol{\omega}}_i + {}^{i}\boldsymbol{\omega}_i \times \dot{\theta}_{i+1}\, {}^{i}\mathbf{p}_{i+1[3\times1]} \right) + \ddot{\theta}_{i+1[1\times1]}\, {}^{i+1}\hat{\mathbf{z}}_{i+1} \tag{6} $$

In this relation, ${}^{i+1}\dot{\boldsymbol{\omega}}_{i+1}$ is the 3×1 angular acceleration vector of link i+1 with respect to the inertial frame, expressed in frame i+1; ${}^{i}\mathbf{p}_{i+1}$ is the 3×1 position vector of the origin of frame i+1 relative to frame i; and $\ddot{\theta}_{i+1}$ is a scalar representing the second derivative of the joint angle $\theta_{i+1}$, corresponding to the angular acceleration about the z-axis of frame i+1. Now, having obtained the angular velocity and angular acceleration of each link, the linear velocity and linear acceleration of each link can be derived through velocity propagation, moving from the first link toward the higher-numbered links. Using the values from (4) to (6), the linear velocity and linear acceleration are then calculated according to (7) and (8), respectively [30].

$$ {}^{i+1}\mathbf{v}_{i+1[3\times1]} = {}^{i+1}\mathbf{R}_i \left( {}^{i}\mathbf{v}_i + {}^{i}\boldsymbol{\omega}_i \times {}^{i}\mathbf{p}_{i+1} \right) \tag{7} $$

$$ {}^{i+1}\dot{\mathbf{v}}_{i+1[3\times1]} = {}^{i+1}\mathbf{R}_i \left( {}^{i}\dot{\boldsymbol{\omega}}_i \times {}^{i}\mathbf{p}_{i+1} + {}^{i}\boldsymbol{\omega}_i \times ({}^{i}\boldsymbol{\omega}_i \times {}^{i}\mathbf{p}_{i+1}) + {}^{i}\dot{\mathbf{v}}_i \right) \tag{8} $$

In these relations, ${}^{i+1}\mathbf{v}_{i+1}$ and ${}^{i+1}\dot{\mathbf{v}}_{i+1}$ are the 3×1 linear velocity and linear acceleration vectors of link i+1 with respect to the inertial frame, expressed in frame i+1. Now, using the values from (4), (6), and (8), the linear acceleration of the center of mass of each link is calculated according to (9) [30].

$$ {}^{i+1}\dot{\mathbf{v}}_{c_{i+1[3\times3]}} = {}^{i+1}\dot{\boldsymbol{\omega}}_i \times {}^{i}\mathbf{p}_{c_{i+1}} + {}^{i+1}\boldsymbol{\omega}_{i+1} \times \left( {}^{i+1}\boldsymbol{\omega}_{i+1} \times {}^{i+1}\mathbf{p}_{c_{i+1}} \right) + {}^{i+1}\dot{\mathbf{v}}_{i+1} \quad (9) $$

In this relation, ${}^{i+1}\dot{\mathbf{v}}_{c_{i+1}}$ is the 3×1 linear acceleration vector of the center of mass of link i+1 with respect to the inertial frame, expressed in frame i+1; and ${}^{i}\mathbf{p}_{c_{i+1}}$ is the 3×1 position vector of the center of mass of link i+1 relative to inertial frame, expressed in frame i.
Since the motion occurs in the x-y plane, the effect of gravity is neglected, as the gravitational force acts on the finger's base and does not influence its lateral motion. Now, having the linear velocity of the center of mass and the angular velocity of each link in their respective local frames, the resultant force and torque acting on each link can be calculated using (4), (6), and (9), according to (10) and (11), respectively [30].

$$ {}^{i+1}\mathbf{f}_{i+1[3\times1]} = m_{i+1[1\times1]} \, {}^{i+1}\dot{\mathbf{v}}_{c_{i+1}} \quad (10) $$

$$ {}^{i+1}\mathbf{n}_{i+1[3\times1]} = {}^{c_{i+1}}\mathbf{I}_{i+1[3\times3]} \, {}^{i+1}\dot{\boldsymbol{\omega}}_{i+1} + {}^{i+1}\boldsymbol{\omega}_{i+1} \times \left( {}^{c_{i+1}}\mathbf{I}_{i+1[3\times3]} \, {}^{i+1}\boldsymbol{\omega}_{i+1} \right) \quad (11) $$

In these relations, ${}^{i+1}\mathbf{f}_{i+1}$ is the 3×1 force vector of link i+1 with respect to the inertial frame, expressed in frame i+1; the scalar $m_{i+1}$ represents the mass of link i+1; ${}^{i+1}\mathbf{n}_{i+1}$ is the 3×1 torque vector of link i+1 with respect to the inertial frame, expressed in frame i+1; and ${}^{c_{i+1}}\mathbf{I}_{i+1}$ is the 3×3 inertia matrix of link i+1, defined with respect to the inertial frame, represented at the center of mass, and expressed in the local coordinate frame of the center of mass. Now, by propagating from the last link toward the base, and substituting (10) and (11), the forces and torques acting on each link are calculated according to (12) to (14). Since all degrees of freedom are rotational, the torque applied to each link is obtained from (14).

$$ {}^{i}\mathbf{g}_{i[3\times1]} = {}^{i}_{i+1}\mathbf{R} \, {}^{i+1}\mathbf{g}_{i+1} + {}^{i}\mathbf{f}_i \quad (12) $$

$$ {}^{i}\mathbf{q}_{i[3\times1]} = {}^{i}\mathbf{n}_i + {}^{i}_{i+1}\mathbf{R} \, {}^{i+1}\mathbf{n}_{i+1} + {}^{i}\mathbf{p}_{c_{i+1}} \times {}^{i}\mathbf{f}_i + {}^{i}\mathbf{p}_{i+1} \times \left( {}^{i}_{i+1}\mathbf{R} \, {}^{i+1}\mathbf{f}_{i+1} \right) \quad (13) $$

$$ \tau_{i[1\times1]} = {}^{i}_{i}\mathbf{q}^{\mathrm{T}} \, {}^{i}\mathbf{z}_i \quad (14) $$

Now, having obtained the torques applied to each link, the dynamic equation resulting from the mass and inertia of the rigid links of the fin ray effect finger can be derived in joint space, as expressed in (15) [30].

$$ \tau_{\mathrm{RB}} = \mathbf{M}(\boldsymbol{\theta})\ddot{\boldsymbol{\theta}} + \mathbf{b}(\boldsymbol{\theta}, \dot{\boldsymbol{\theta}}) \quad (15) $$

In this study, the finger's components are modeled using torsional spring–damper couplings to characterize their mechanical properties and behavior. As motion-induced accelerations are negligible, inertial effects (mass and rotational inertia) are disregarded, and quasi-static equilibrium equations are derived for the gripper finger. These equations provide preliminary estimates of torsional stiffness and damping characteristics, supported by complementary low-speed experimental tests. The quasi-static formulation neglects dynamic effects and considers only force and moment equilibrium. Governing moments comprise the actuator torque applied at the finger base, the contact-induced moment between the finger and the object applied to the fifth link of the finger, and the elastic and viscous moments generated at the joints. As confirmed experimentally, these constant forces retain their direction during finger operation (Fig. 6).

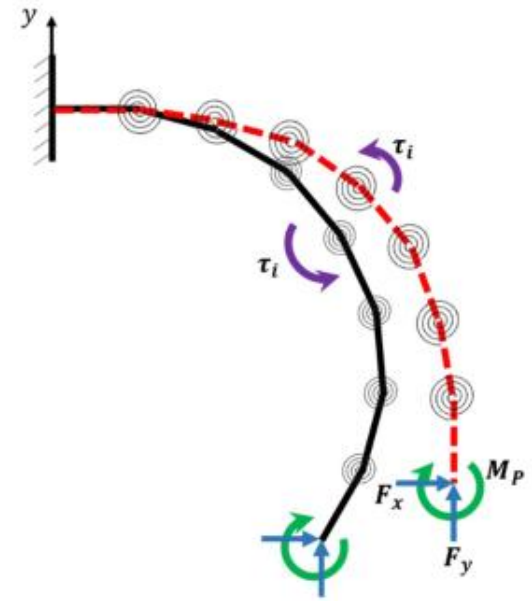

**Fig. 6.** Constant-direction force [28].

When the constant external force applied to the fifth link has horizontal and vertical components, $F_x$ and $F_y$, the corresponding force vector is defined as $F = [F_x . F_y . 0 . 0 . 0 . M]^T$. Under these equilibrium conditions, the joint torques can be determined using (16); in this equation $\theta'_t$ is defined as $\theta'_t = \theta_1 + \theta_2 + \ldots + \theta_t$ [28].

$$ \tau_i = {}^{\mathbf{0}}\mathbf{J}_i^T \mathbf{F} = \left( -\sum_{t=0}^{n-1} a_t \times \sin\theta'_t + \sum_{t=0}^{i-1} a_t \times \sin\theta'_t \right) F_x + \left( \sum_{t=0}^{n-1} a_t \times \cos\theta'_t - \sum_{t=0}^{i-1} a_t \times \cos\theta'_t \right) F_y + M \quad (16) $$

Additionally, the reaction torques resulting from the torsional springs and dampers at the joints are defined for the i-th link as functions of stiffness and damping coefficients, in accordance with (17).

$$ \tau_i = k_i\theta_i + c_i\dot{\theta}_i \quad (17) $$

As a result, the combination of (16) and (17) leads to the derivation of the final equations, according to (18), which can be used to estimate the stiffness and damping of the links.

$$ k_i\theta_i + c_i\dot{\theta}_i + F_x \sum_{t=i}^{n-1} a_t \sin\theta'_t - F_y \sum_{t=i}^{n-1} a_t \cos\theta'_t = M, \quad i = 1, 2, \ldots, n-1 \quad (18) $$

It is assumed that the stiffness and damping of the rigid elements near the base of the gripper finger are greater than those near the fingertip. Accordingly, the stiffness and damping of the first link are considered as $k_1$ and $c_1$, respectively, and the stiffness and damping of the remaining links are calculated based on a parabolic relation, dependent on $k_1$ and $c_1$, in accordance with (19), where N is the number of links and $k_1$ is the stiffness of the first link. The damping of each link is distributed in a similar manner.

$$ k_i = k_1 \left( 1 - \frac{i-1}{N-1} \right)^2, \quad i = 1, 2, \ldots, n-1 \quad (19) $$

By rewriting (18) in accordance with (19), (20) is obtained [28], where $s_i = \sin\theta'_i$ and $c_i = \cos\theta'_i$ are defined.

$$\begin{cases} k_1\theta_1 + c_1\dot{\theta}_1 + F_x\left(a_4s_4 + a_3s_3 + a_2s_2 + a_1s_1\right) - F_y dl\left(a_4c_4 + a_3c_3 + a_2c_2 + a_1c_1\right) = M, \\ \left(\frac{6}{7}\right)^2 k_1\theta_2 + \left(\frac{6}{7}\right)^2 c_1\dot{\theta}_2 + F_x\left(a_4s_4 + a_3s_3 + a_2s_2\right) - F_y\left(a_4c_4 + a_3c_3 + a_2c_2\right) = M, \\ \left(\frac{5}{7}\right)^2 k_1\theta_3 + \left(\frac{5}{7}\right)^2 c_1\dot{\theta}_3 + F_x\left(a_4s_4 + a_3s_3\right) - F_y dl\left(a_4c_4 + a_3c_3\right) = M, \\ \left(\frac{4}{7}\right)^2 k_1\theta_4 + \left(\frac{4}{7}\right)^2 c_1\dot{\theta}_4 + F_x\left(a_4s_4\right) - F_y dl\left(a_4c_4\right) = M. \end{cases}$$

(20)

A controlled moment is applied at the robot base, and the resulting contact force on the fifth link is measured with a load cell, while link angles are extracted via image processing. The measured moment, contact force, and link angles (TABLE IV) are substituted into the quasi-static equation (20) to estimate the stiffness and damping of the first link. The corresponding parameters of the remaining links are then obtained using (19). The resulting stiffness and damping across the links are presented in TABLE V.

Table IV
THE MOMENT, CONTACT FORCE, AND LINK ANGLES OBTAINED FROM THE EXPERIMENTAL TEST

| Moment (N.m) | Contact Force (N) | $\theta_1$ (rad) | $\theta_2$ (rad) | $\theta_3$ (rad) |
|---|---|---|---|---|
| 0.49 | 5.6 | -0.203 | -0.106 | -0.041 |
| **$\theta_4$ (rad)** | **$\theta_5$ (rad)** | **$\theta_6$ (rad)** | **$\theta_7$ (rad)** | **$\theta_8$ (rad)** |
| 0.079 | 0.207 | 0.302 | 0.111 | -0.026 |

Table V
THE STIFFNESS AND DAMPING VALUES DETERMINED FOR THE RIGID ELEMENTS OF THE GRIPPER FINGER

| Link i | 1 | 2 | 3 | 4 | 5 | 6 | 7 |
|---|---|---|---|---|---|---|---|
| Stiffness $[\frac{Nm}{rad}]$ | 0.897 | 0.659 | 0.458 | 0.293 | 0.165 | 0.073 | 0.018 |
| Damping $[\frac{Nms}{rad}]$ | 0.090 | 0.066 | 0.046 | 0.029 | 0.017 | 0.007 | 0.002 |

Finally, the derived stiffness and damping model is compared with the experimental data for validation. This methodology, combining experimental data with precise modeling, provides a comprehensive understanding of the mechanical properties of the system, and the parabolic stiffness distribution accurately reflects the system's physical characteristics. Now, it is necessary to calculate the torques of the dynamic couplings. To compute the torques resulting from the torsional springs and dampers, (21) and (22) are used, respectively [30].

$$\boldsymbol{\tau}_{\mathbf{k}[8\times1]} = \begin{bmatrix} k_1 & 0 & 0 & \cdots & 0 \\ 0 & k_2 & 0 & \cdots & 0 \\ \vdots & \vdots & \vdots & \ddots & \vdots \\ 0 & 0 & k_{i[i,i]} & \cdots & 0 \\ \vdots & \vdots & \vdots & \ddots & \vdots \\ 0 & 0 & 0 & \cdots & k_8 \end{bmatrix}_{8\times8} \boldsymbol{\theta}_{[8\times1]} \tag{21}$$

$$\boldsymbol{\tau}_{\mathbf{C}[8\times1]} = \begin{bmatrix} c_1 & 0 & 0 & \cdots & 0 \\ 0 & c_2 & 0 & \cdots & 0 \\ \vdots & \vdots & \vdots & \ddots & \vdots \\ 0 & 0 & c_{i[i,i]} & \cdots & 0 \\ \vdots & \vdots & \vdots & \ddots & \vdots \\ 0 & 0 & 0 & \cdots & c_8 \end{bmatrix}_{8\times8} \dot{\boldsymbol{\theta}}_{[8\times1]} \tag{22}$$

To calculate the external torque resulting from the contact force, it is assumed that the contact force acting on the gripper finger always remains perpendicular to the finger's central curve. To simplify the equations, the external contact force is assumed to act as a point load at the center of mass of each degree of freedom, as illustrated in Fig. 7. To reduce the error introduced by representing a distributed force as a point load, the number of degrees of freedom can be increased if necessary. In the experimental tests, the contact between the finger and the load cell was considered as a point load with a fixed direction. Therefore, in system modeling, the relations corresponding to a constant-direction force are used. This approach is adopted because the results of theoretical modeling are ultimately validated against the actual system data, making it essential to align the experimental conditions with the mathematical model. For this purpose, it is necessary to extract the Jacobian corresponding to the fifth degree of freedom, where the finger's contact point with the load cell is located. The Jacobian of the fifth degree of freedom with respect to the inertial coordinates can be calculated using (23) [28].

$$^{0}\mathbf{J}_{5\,[6\times8]} = \begin{cases} \begin{bmatrix} ^{0}\mathbf{z}_j \times \left(^{0}o_5 - ^{0}o_j\right) \\ ^{0}\mathbf{z}_j \end{bmatrix} & \text{if } j < 5 \\ zeros(6,1) & \text{if } j \geq 5 \end{cases} \qquad j = 1,2,\ldots,n \tag{23}$$

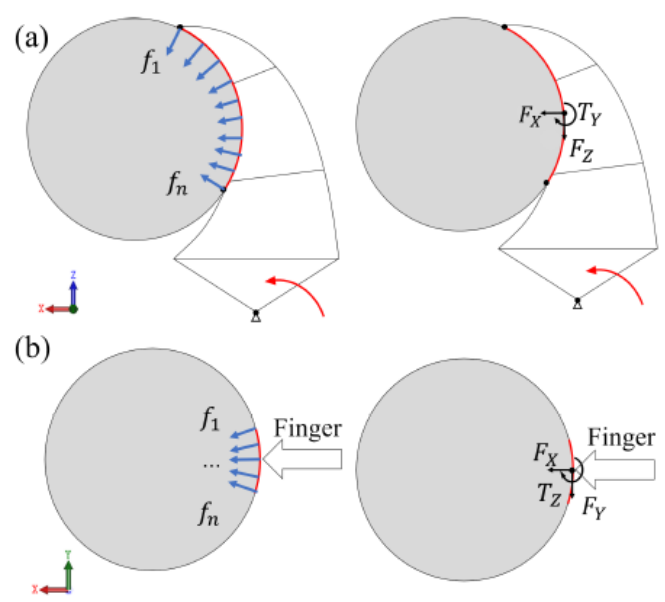

**Fig. 7.** Schematic of single-finger with a distributed force in (a) the X–Z plane (b) the X–Y plane [31].

Now, by extracting the Jacobian matrix, to compute the torque resulting from the contact force, the torque at the center of mass of each degree of freedom must be calculated, according to (24) to (26), where $\mathbf{f}_{c[6\times1]}$ is the external force vector resulting from the contact force [28].

$$\boldsymbol{\tau}_{f\,[8\times1]} = \sum \langle i-j \rangle {}^{i}\mathbf{J}_i^{\bullet}\, \mathbf{f}_{c[6\times1]} \tag{24}$$

$$\langle i-j \rangle = \begin{cases} 1 & \text{if } j \leq i, \\ 0 & \text{if } j > i. \end{cases} \tag{25}$$

$$\mathbf{f}_{c[6\times1]} = \left[0; -F; 0; 0; 0; 0\right] \tag{26}$$

Craig's robotic formulations are combined with the rigid finite element model to derive the governing equations of the soft gripper finger. The formulation incorporates torques from system behavior (15), dynamic couplings (21) and (22), external contact force (24), and the motor-applied moment at the finger base. The resulting equations, expressed in the form of (27), are based on Newton's second law and equilibrium relations and are prepared for implementation in solver software [30].

$$\tau_{\text{RB}} + \tau_{\text{k}} + \tau_{\text{c}} = \tau_{\text{Motor}} - \tau_{f} \tag{27}$$

### *C. Experimental Setup and Data Acquisition*

In this section, the methodology for validating the derived dynamic relations of the finger is presented. The experimental setup is first introduced, including its components and design considerations. The robotic gripper finger, fabricated from TPU95A using 3D printing, and based on a fish-fin mechanism, is mounted on the testbed, and data are collected for model verification, as illustrated in Fig. 8. Motion data of the finger is obtained via video recordings processed using image analysis, while force data are measured using a uniaxial bending load cell. The finger, made of soft material, is connected to the motor and encoder via a PLA joint, providing sufficient strength and space for mounting sensors, electronics, and a camera.

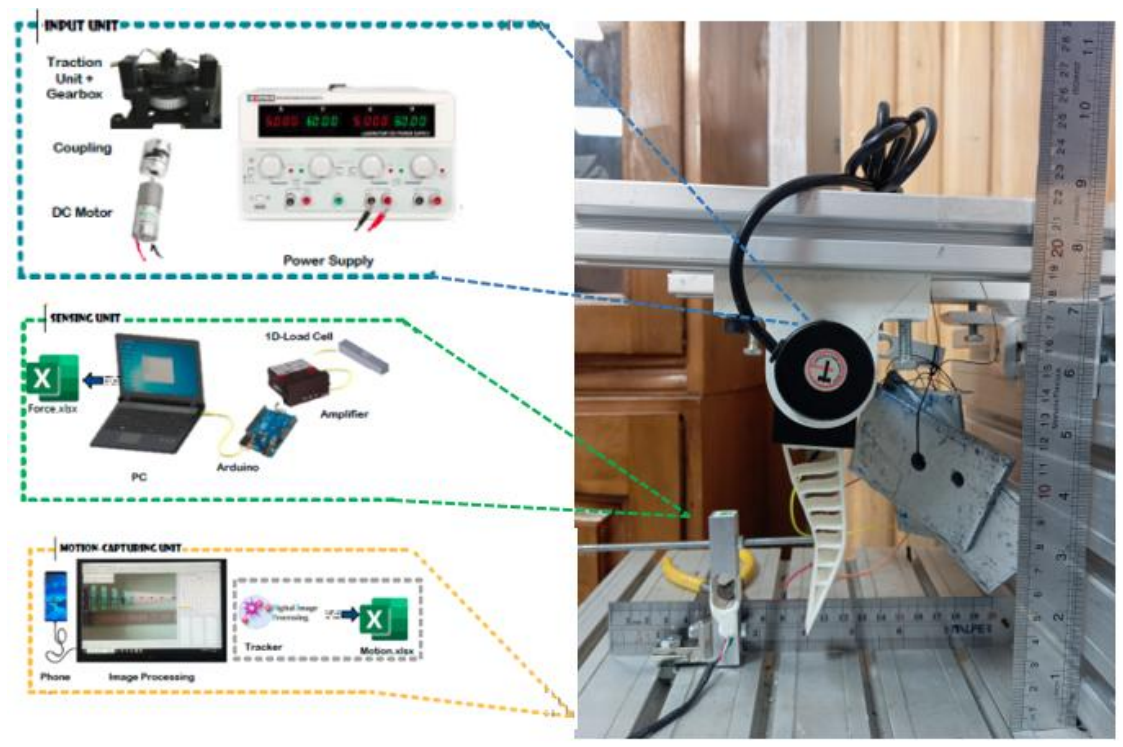

**Fig. 8.** Conceptual diagram of the testbed

To measure system input, the motor applies torque, and the corresponding current is recorded via an INA219 sensor and an Arduino microcontroller. This current-to-torque relation is calibrated and used to compute the motor input torque. Simultaneously, finger bending is captured through video at 60 frames per second and processed offline using Tracker software, where markers placed on the finger track positional changes over time. Contact forces are measured using the load cell mounted perpendicularly at the base of the finger, and the sensor output voltage is converted to force based on strain gauge principles.

The testbed design ensures precise alignment of sensors, motor, and finger components to replicate experimental conditions consistently. The motor torque calibration involves applying known loads via weights and measuring the resulting current, which allows extraction of the motor stiffness coefficient. This process ensures that motor inputs in dynamic equations accurately reflect experimental conditions. The motor torque calibration system is shown on Fig. 9.

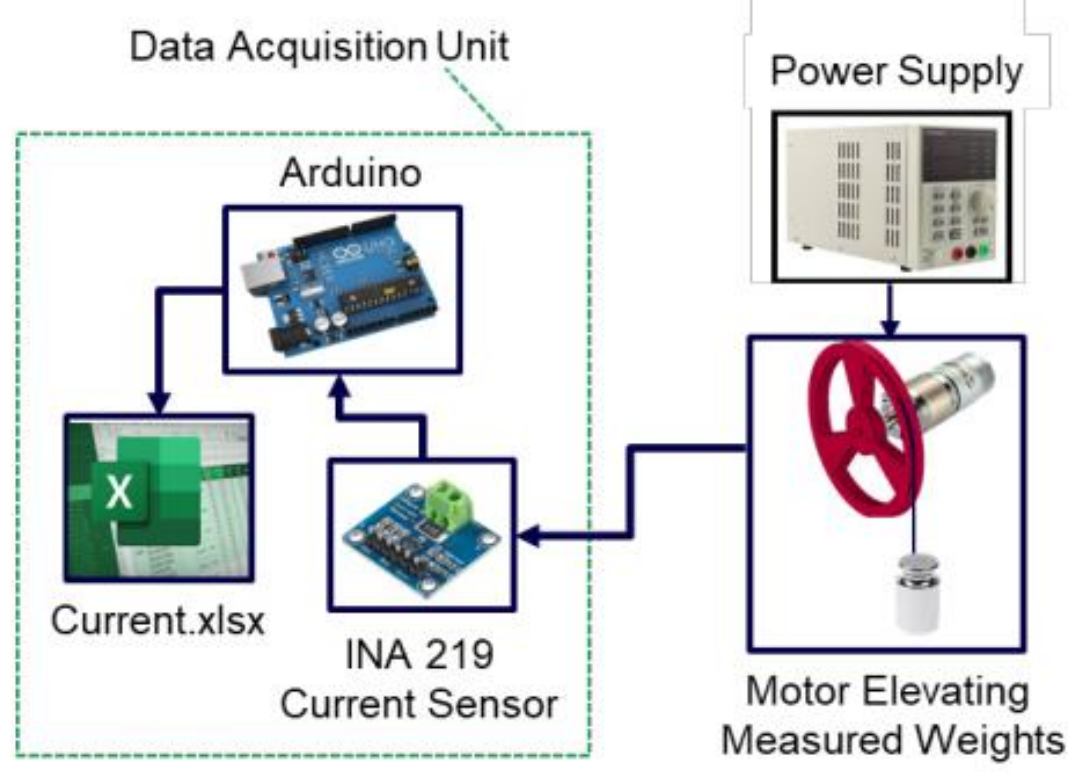

**Fig. 9.** Procedure for extracting the motor stiffness coefficient

Once all elements are properly installed, the testbed enables recording of motion, contact force, and input torque data during finger operation. These measurements provide the necessary information for validating the dynamic model, ensuring alignment between theoretical predictions and the physical behavior of the system.

## III. Results and Discussion

### *A. Optimized design of the gripper finger*

This section presents the performance evaluation of the Fin Ray-inspired gripper finger, carried out by displacing an object 20 mm into the structure to identify optimal geometric parameters, including finger width, rib spacing, rib angle, and rib thickness. The analysis focused on X-direction displacement (adaptability to object position), Y-direction displacement (enclosure stability), contact force (load-bearing capacity), and stress (structural durability). For finger width, a value of 30 mm (Fig. 10) provided the best balance, with fingertip displacement in the X-direction reaching 2.36 mm and Y-direction displacement limited to 0.13 mm, ensuring adaptability and precision. Total deformation was minimized at 20.07 mm, stress remained favorable at 3.39 MPa, and the maximum contact force reached 3.57 N.

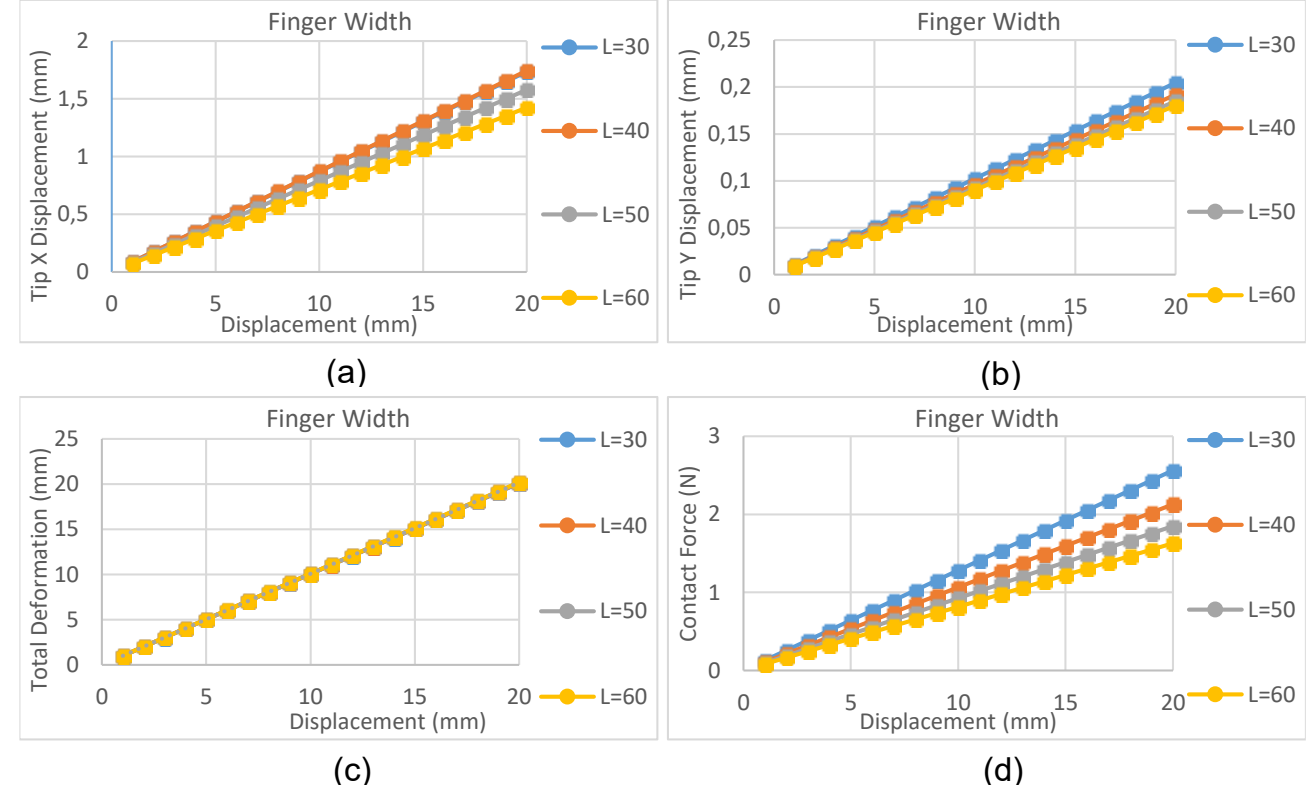

**Fig. 10.** (a) Fingertip displacement (X), (b) Fingertip displacement (Y), (c) Overall deformation, (d) Stress, and (e) Contact force versus object displacement at different finger widths.

For rib spacing, 10 mm was identified as optimal (Fig. 11), offering sufficient X-direction displacement (2.15 mm) for adaptability while Y-displacement remained low (0.11 mm) for stable enclosure. Total deformation (20.12 mm), stress (3.25 MPa), and contact force (3.45 N) confirmed an effective balance between stiffness, flexibility, and load-bearing capacity.

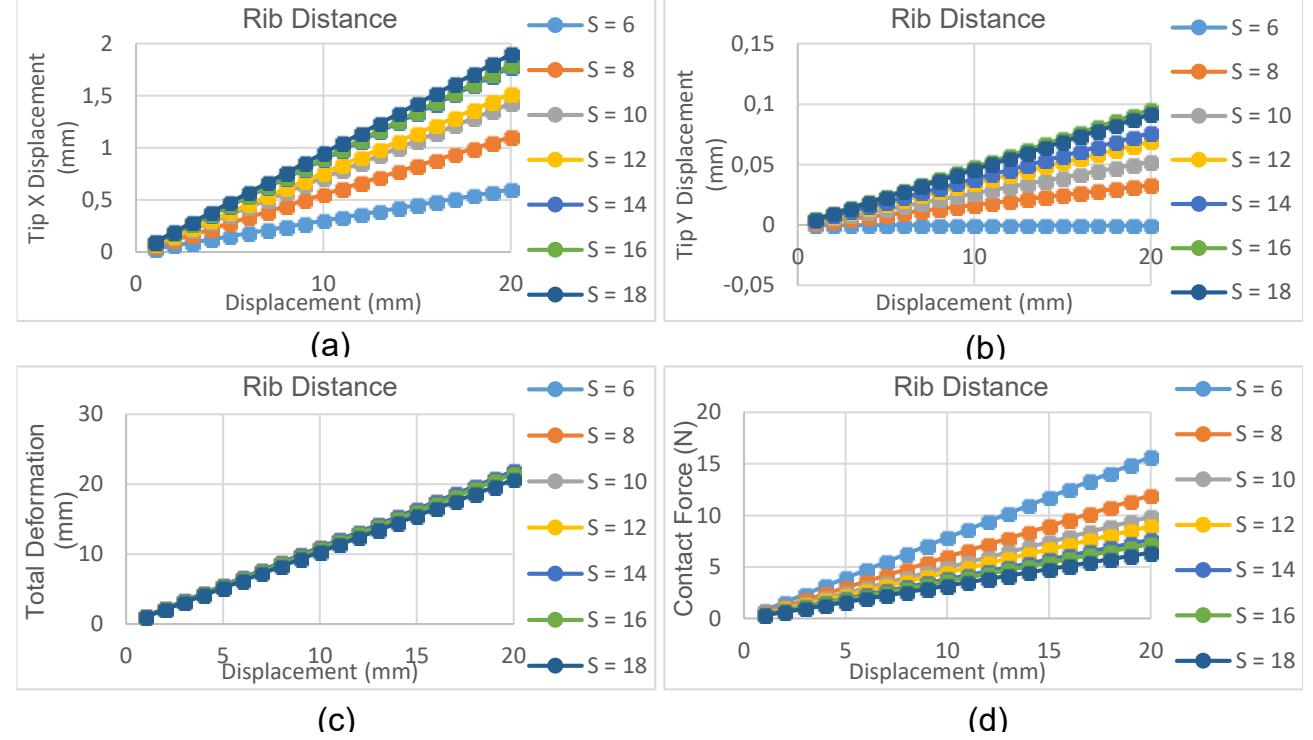

**Fig. 11.** (a) Fingertip displacement (X), (b) Fingertip displacement (Y), (c) Overall deformation, (d) Stress, and (e) Contact force versus object displacement at different rib distances.

Rib angle was optimized at –15° (Fig. 12), where X- and Y-direction displacements were 1.62 mm and 0.072 mm, respectively, reflecting good adaptability and precise enclosure. Total deformation was 20.5 mm, stress decreased to 3.2 MPa, and the maximum contact force increased to 8.9 N, demonstrating enhanced structural stability and gripping strength.

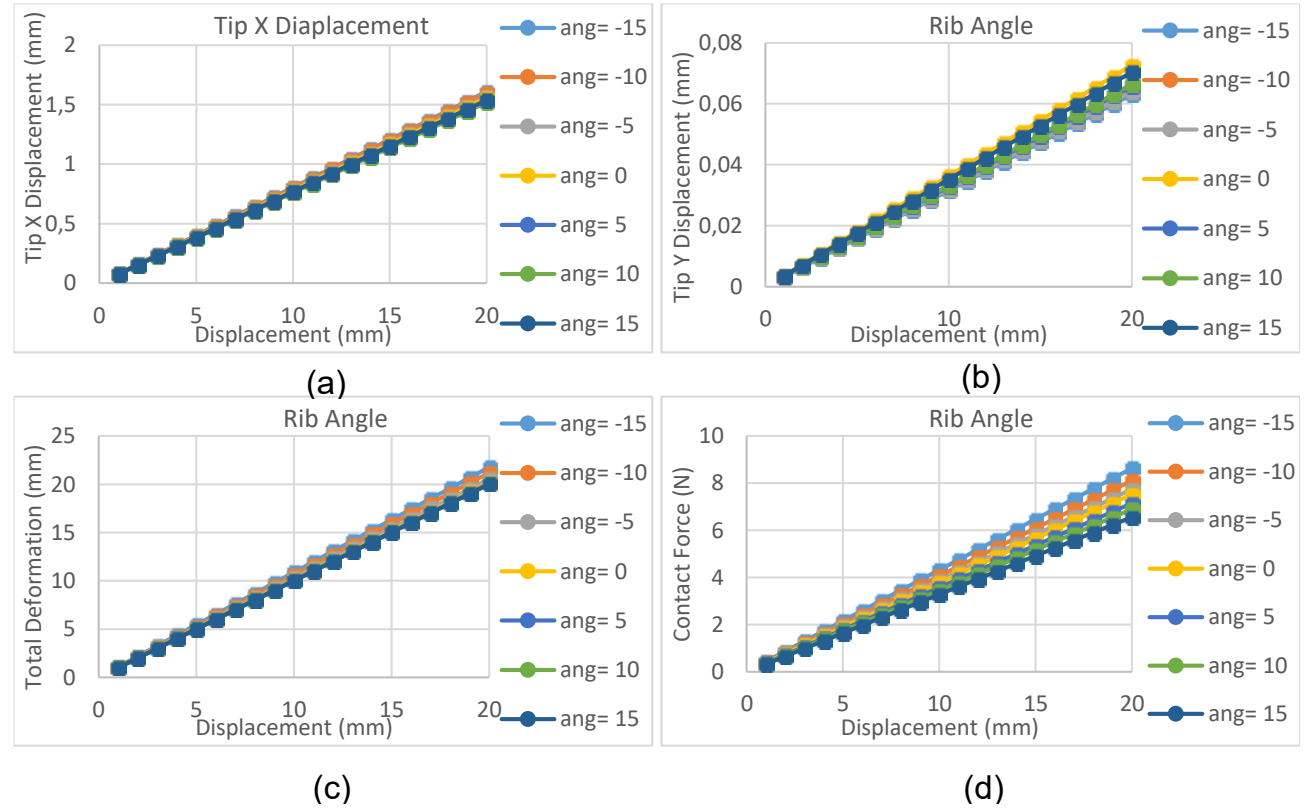

**Fig. 12.** (a) Fingertip displacement (X), (b) Fingertip displacement (Y), (c) Overall deformation, (d) Stress, and (e) Contact force versus object displacement at different rib angles.

Finally, rib thickness of 1 mm (Fig. 13) achieved optimal performance, with maximal X-direction displacement, minimal Y-displacement, balanced total deformation, favorable stress levels, and a maximum contact force confirming superior load capacity and grip strength. Overall, these results indicate that the identified optimal parameters, 30 mm width, 10 mm rib spacing, –15° rib angle, and 1 mm rib thickness, provide a well-balanced combination of flexibility, stiffness, stability, and durability, ensuring efficient and reliable performance of the gripper finger for handling delicate objects. Under this configuration, the simulation with object displacement confirmed robust performance: X- and Y-displacements of 1.4253 mm and 0.0522 mm, total deformation of 21.858 mm, stress of 4.2071 MPa, and a maximum contact force of 9.884 N.

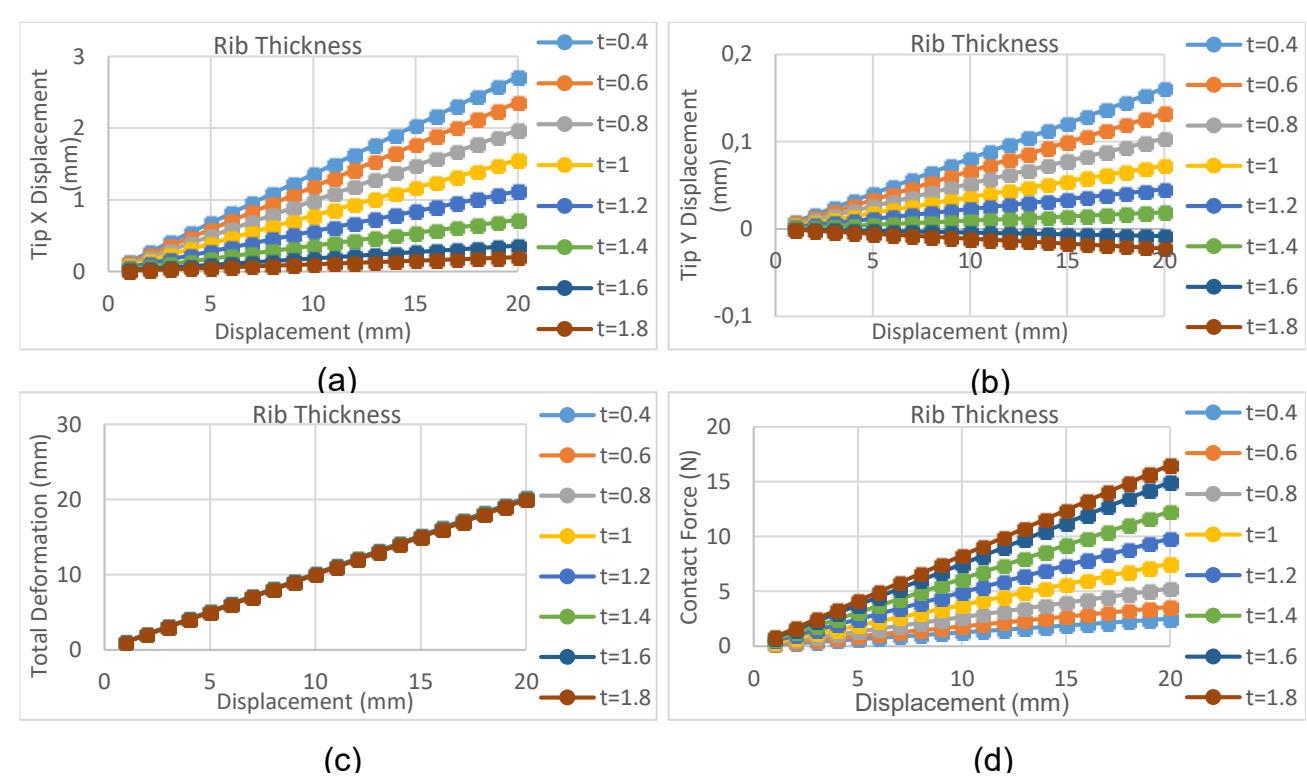

**Fig. 13.** (a) Fingertip displacement (X), (b) Fingertip displacement (Y), (c) Overall deformation, (d) Stress, and (e) Contact force versus object displacement at different rib thicknesses.

The observed trends in the simulations can be explained by the mechanical behavior of the gripper's Fin Ray-inspired structure. Increased width and rib thickness improve stiffness and load-bearing capacity, while optimal rib spacing and angle enhance adaptability and precise enclosure. Specifically, the –15° rib angle promotes effective bending along the object's surface, and a 1 mm rib thickness provides sufficient flexibility without compromising structural integrity. Together, these geometric features enable the gripper to conform to objects of varying shapes and sizes, distribute contact forces uniformly, and maintain durability under repeated use.

*B. Dynamic Modeling and Experimental Validation*

The dynamic behavior of the bio-inspired gripper was analyzed using three complementary approaches: the Finite Rigid Elements Method for theoretical modeling, nonlinear finite element simulations in Ansys Workbench, and experimental testing. FREM was employed to derive the governing equations of finger bending, while simulations captured force distribution, moments, and deformations under realistic conditions. Experimental results validated both models, confirming their accuracy and applicability for force-control design. This integrated validation establishes a reliable foundation for precise and optimized gripper control and performance.

The influence of contact force on gripper finger bending was investigated using FREM. Four scenarios with applied forces of 4, 6, 8, and 10 N on the fifth link were simulated, and the resulting displacement curves are shown in Fig. 14.

Results indicate that bending increases significantly with higher forces, shifting from an almost linear response at low loads (4 N) to a nonlinear trend at higher loads (10 N). At lower forces, deformation is more homogeneous across the structure, while higher forces amplify differences between the upper and lower edges, reflecting the onset of nonlinear effects. Despite minor asymmetries, overall bending remains stable and symmetric, demonstrating the finger's ability to adapt to varying contact forces without mechanical instability. Such behavior is advantageous for soft robotic fingers in tasks requiring safe interaction with fragile objects.

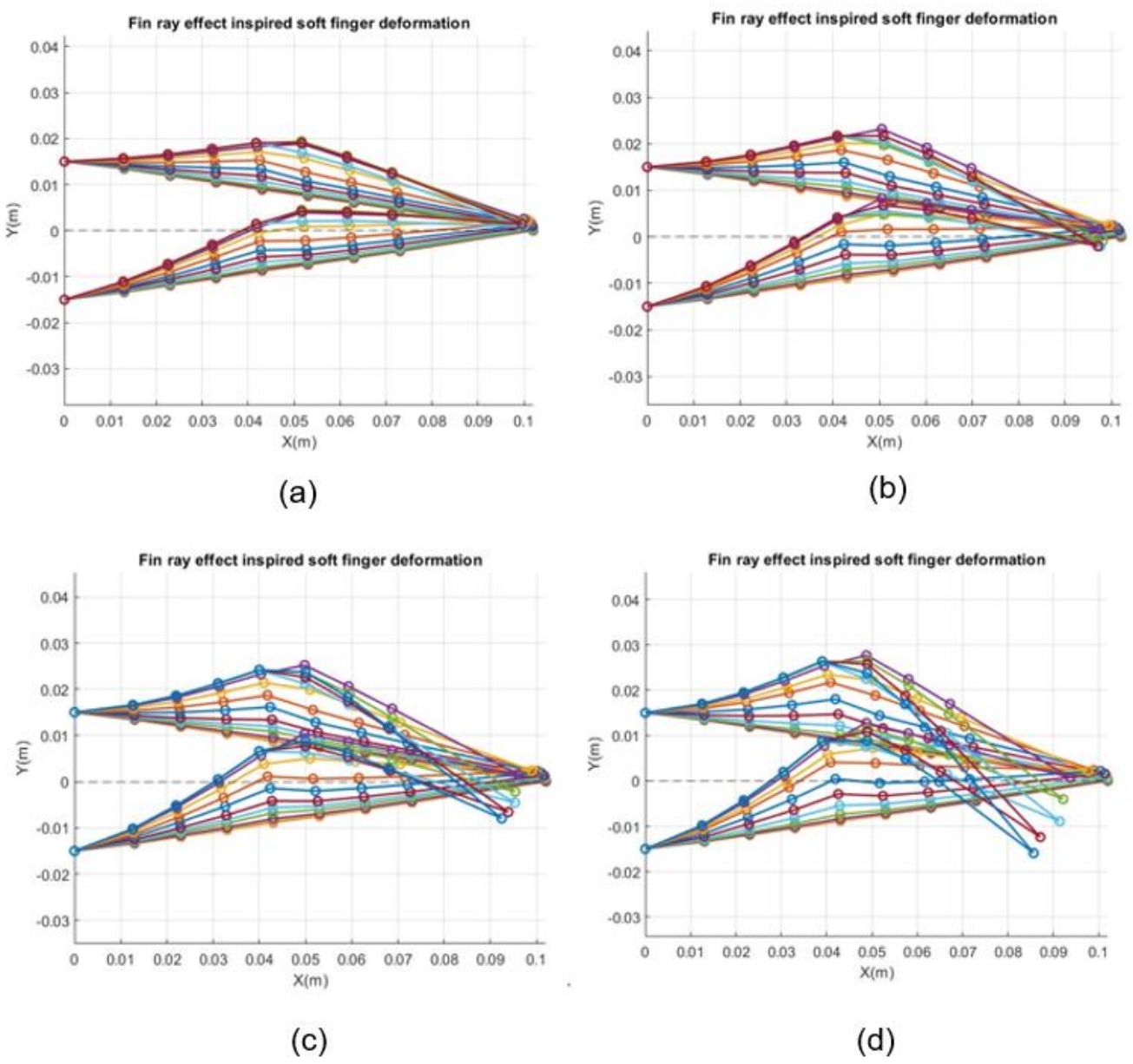

**Fig. 14.** Time-dependent bending response of the gripper finger under different contact forces: (a) 4 N, (b) 6 N, (c) 8 N, (d) 10 N.

The gripper finger performance was evaluated using three complementary approaches: experimental testing, finite element simulations in Ansys Workbench, and the FREM theoretical model implemented in MATLAB. In all cases, a torque of 0.4944 $N \cdot m$ was applied at the finger base, resulting in a measured contact force of 5.5952 N.

Experimental tests (Fig. 15) showed that the finger undergoes significant bending, with maximum displacement occurring in the mid-span region and reduced deformation near the base. These results highlight the compliant behavior of the structure under loading.

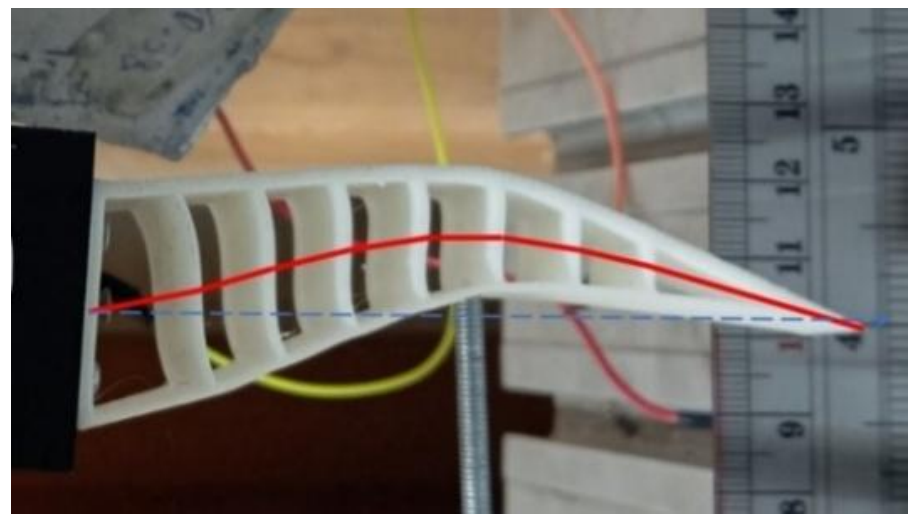

**Fig. 15.** Experimental bending of the gripper finger under applied torque and contact force

Finite element simulations in Ansys (Fig. 16) reproduced the same deformation trend, showing maximum stress at the joint and maximum displacement at the mid-span.

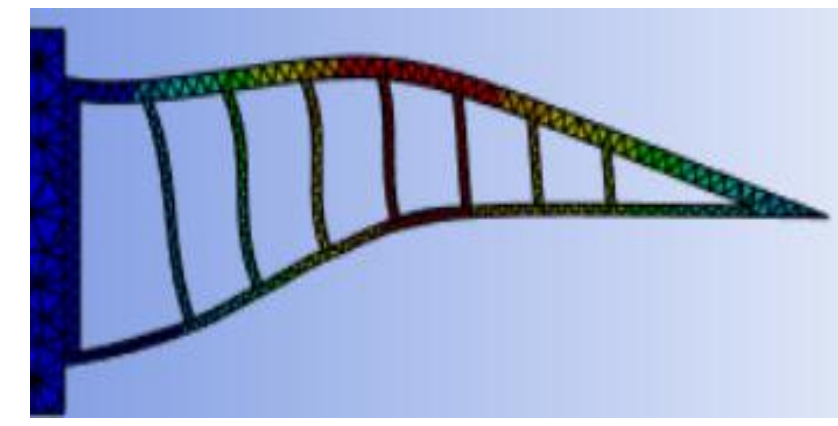

**Fig. 16.** Simulated gripper finger bending in Ansys.

However, comparison with experiments (Fig. 17) revealed that Ansys slightly underestimated bending. This discrepancy is attributed to fabrication tolerances, variations in material properties, and unmodeled boundary effects such as joint compliance and friction.

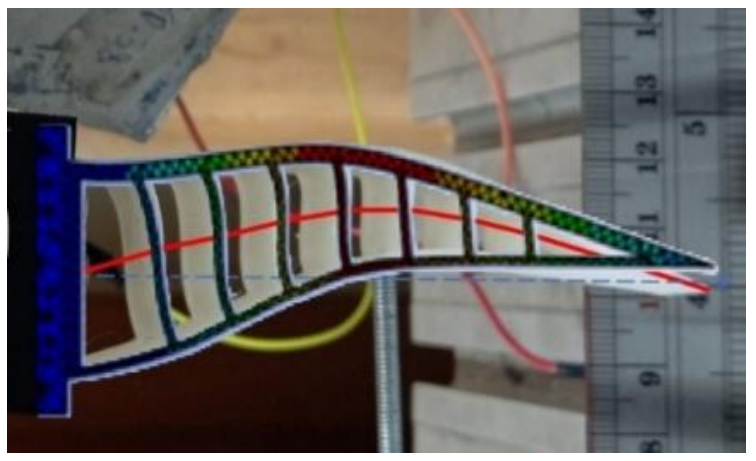

**Fig. 17.** Comparison with experimental bending for FEM validation

The FREM theoretical model (Fig. 18) captured the overall deformation trend but consistently predicted smaller displacements compared with experimental observations. This underestimation results from simplifying assumptions that neglect detailed stress and strain distributions.

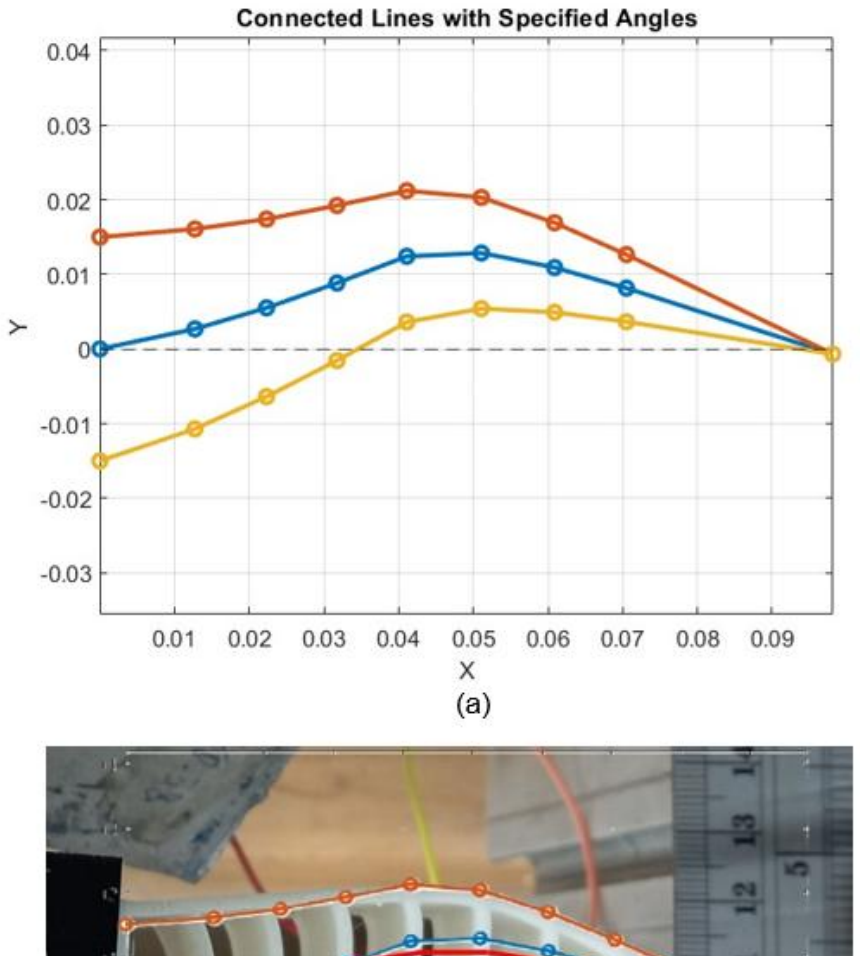

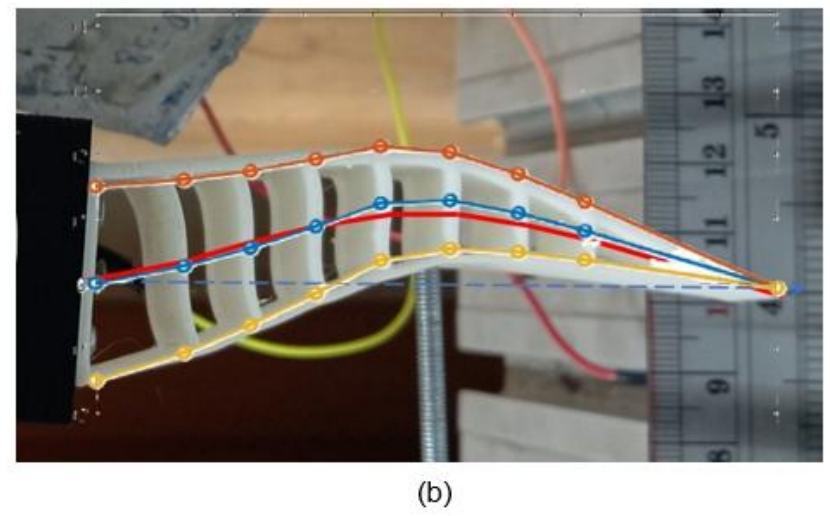

(b)

**Fig. 18.** (a) Gripper finger bending from the FREM model. (b) Comparison with experimental bending for model validation.

Angle deviation analysis of the gripper finger links was performed using the theoretical FREM model, finite element simulations, and experimental tests. The FREM model captured the overall trend of link rotations, with discrepancies from experimental data within acceptable limits. These differences are primarily due to model simplifications, including assumptions on link stiffness, neglecting joint compliance, and simplified stress distribution. FEM simulations showed higher accuracy, with smaller deviations from experimental results, reflecting their superior ability to capture detailed stress and strain distributions.

Table 6 summarizes the final link angles and prediction errors for both theoretical and numerical methods. Both approaches provide reliable estimates of finger bending under the applied load. While FEM is more precise, the FREM model remains a computationally efficient and sufficiently accurate tool for preliminary analysis and can serve as a basis for force-controller design.

Table VI
LINK ANGLES AND PREDICTION ERRORS OF THEORETICAL AND NUMERICAL MODEL

| Link | 1 | 2 | 3 | 4 | 5 | 6 | 7 | 8 |
|---|---|---|---|---|---|---|---|---|
| **Experiment** | 0.205 | 0.08 | 0.052 | 0.026 | -0.315 | -0.237 | -0.083 | -0.03 |
| **Theory** | 0.209 | 0.079 | 0.052 | 0.026 | -0.318 | -0.242 | -0.085 | -0.03 |
| **Simulation** | 0.208 | 0.079 | 0.052 | 0.026 | -0.321 | -0.241 | -0.084 | -0.03 |
| **Theoretical Error** | 1.963 % | 1.067 % | 1.581 % | 0.992 % | 0.959 % | 2.158 % | 1.663 % | 0.129 % |
| **Numerical Error** | 1.686 % | 0.808 % | 1.557 % | 0.252 % | 1.832 % | 1.347 % | 0.871 % | 0.759 % |

## IV. CONCLUSION

This study presented the design, optimization, and dynamic modeling of a fish-fin-inspired soft robotic gripper, focusing on achieving an optimal balance between stiffness and flexibility. Parametric simulations in ANSYS Workbench were used to evaluate finger performance under varying geometrical parameters, including bending, tip displacements, stress, and contact force. The analyses demonstrated that increased finger bending improves adaptability to object positioning, while appropriate tip displacements and reduced stress enhance grasp stability and gripper lifespan.

The optimized finger dimensions, 30 mm width, 10 mm rib spacing, -15° rib angle, and 1 mm rib thickness, provided the best compromise between precision, force efficiency, durability, and structural robustness. Dynamic modeling using the FREM in MATLAB accurately captured overall finger deformation, enabling efficient prediction of contact forces and displacements for controller design. While FEM simulations in ANSYS offered higher accuracy in stress and strain predictions, their computational cost limits real-time applications. Combining these approaches allows rapid theoretical modeling for control and high-fidelity numerical validation for design optimization. This framework demonstrates that bio-inspired simulation-based design can effectively guide the development of high-performance, soft robotic grippers and serves as a foundation for future implementation in intelligent, real-time control systems.